\documentclass[runningheads]{llncs}
\usepackage{graphicx}
\usepackage{makecell}
\usepackage{multirow}

\usepackage{esvect}

\usepackage[tight,TABTOPCAP]{subfigure}
\usepackage{cite}
\usepackage{booktabs}
\usepackage{stfloats}
\usepackage{amsmath}
\begin{document}
%
\title{Balancing Real and Synthetic Data for CNN-based Masonry Crack Detection\\
\large A Preprint\thanks{Preprint. This paper has been accepted for presentation at ARTISTE 2025.}}
%
%
\author{Mattia Forlesi\inst{1}\orcidID{0009--0005--9365--6168} \and
Alfonso Esposito\inst{1}\orcidID{0009--0006--6381--1187} \and
Ivan Zyrianoff\inst{1}\orcidID{0000--0003--4936--9645} \and
Alessandro Marzani\inst{2}\orcidID{0000--0001--7697--6729} \and
Marco Di Felice\inst{1}\orcidID{0000--0001--7496--7597}}
\authorrunning{M. Forlesi et al.}
%
\institute{Department of Computer Science and Engineering, University of Bologna, Bologna, Italy \and
Department of Civil, Chemical, Environmental, and Materials Engineering, University of Bologna, Bologna, Italy\\
\email{\{mattia.forlesi2, alfonso.esposito6, ivandimitry.ribeiro, alessandro.marzani, marco.difelice3\}@unibo.it}}
\maketitle              
\begin{abstract}
Cracks are a critical indicator of building health, and early stage identification is fundamental to prevent harmful damages. Advances in deep learning (DL), particularly convolutional neural networks (CNNs), have enabled scalable solutions for automated crack detection. However, CNN performance strongly depends on the availability of large and diverse datasets, which is particularly challenging for complex surfaces such as masonry. 
Collecting sufficient real data is time-consuming, while publicly available datasets may not be adequate. 
To address this limitation, we explored generating synthetic crack data, which complements real data and improves training effectiveness. The real dataset consists of masonry crack images collected from buildings in Bologna and surrounding areas. In contrast, the synthetic dataset was generated using a crack overlay tool that adds cracks to background images in a controlled orientation and placement. The real dataset was used to train several DL architectures, to identify the best-performing model (InceptionV4) employed for experiments with generated data. Six training scenarios were tested in InceptionV4 by varying the ratio of real and synthetic data, with evaluation performed on a test set composed of real images using the F1-score and mean Intersection over Union (mIoU) metrics. Results show that training on synthetic data plus a modest addition of 20\% real data achieves results comparable to training on real data only. Moreover, the 20/80 scenario (synthetic/real) achieved an 76\% F1-score and 80\% mean IoU, outperforming the real-only case. As can be seen, the method demonstrates the potential of synthetic data to reduce collection efforts while enhancing crack detection accuracy. 

\end{abstract}
\section{Introduction}\label{sez:intro}

The aging and the vulnerability to damage buildings and civil infrastructures has driven to the adoption of Structural Health Monitoring (SHM) systems \cite{digital_zyrianoff_2024}, along with the management of numerous heterogeneous measurements and their complex relationship to structural and environmental behavior \cite{iot_forlesi_2024}. 
Among the several structural features that can be measured and monitored, cracks represent a critical indicator of building health, and 
accurate identification is essential for timely repairs, cost savings, and accident prevention \cite{crack_marsh_2025, adaptive_tan_2024, generative_kim_2024}. Currently, the most traditional methods for crack detection 
can be affected by human errors (such as visual inspection and sensors installation) \cite{syncrack_garcia_2022} and allow a level of monitoring and implementation that is limited to what is assumed in the design stage \cite{monitoring_bacco_2020}. 
The overall diffusion of DL techniques has led to feasible and scalable solutions
improving the capability to detect cracks and provide attribute information through semantic segmentation, a computer vision task aiming to separate different classes at the pixel level to create a pixel-level understanding of the image \cite{generative_kim_2024}, \cite{automatic_dais_2021}. However, the CNN models involved are really dependent on the amount of data necessary for the training phase, where real data collection is often time-consuming, while the datasets available online could be unsatisfactory for the studies to achieve \cite{improving_dondi_2025}, \cite{innovative_xu_2023}.
In this scenario, data augmentation can be considered, enhancing the real image datasets with synthetic crack images. This enables the creation of a more consistent dataset, minimizing the time required for data collection and labeling, while overall enhancing the performance metrics used \cite{enhancing_mahmood_2025}. 
In generative adversial networks (GANs), texture details and fidelity increase sharply and their intrinsic design reduces the experimental time, pushing the model towards a more efficient annotation task \cite{data_li_2024}. However, GANs generate random cracks without any domain-expert supervision logic, often leading to unreliable crack representation in respect to the background image structure 
\cite{improving_dondi_2025}. Moreover, they still require an important amount of images to generate high-quality data samples \cite{crack_marsh_2025}. Hence, an alternative solution is determined by the direct modeling and rendering of crack in 3D computer graphic software, producing more contextualized outcomes with meta-annotations constraints to govern random generation \cite{improving_dondi_2025}. However, cracks generated in this way could lack realistic aesthetic refinement \cite{synthetic_zhai_2022} and the overall modeling operation could be difficult, since it requires computer graphic domain experts \cite{innovative_xu_2023}. In this paper, we propose a method for generating synthetic crack images to enrich small real dataset and train U-Net models (popular CNN models commonly used in literature for segmentation tasks) to evaluate the enhancements in terms of F1-score and mIoU.
Towards this goal, our contributions can be summarized as:
\begin{itemize}
    \item The real dataset was totally collected and labeled by our team. In particular, we focused on the collection of masonry crack images due to the complexity of the material that can test the efficiency of the DL algorithms adopted.
    \item We proposed a custom image processing method for the generation of synthetic cracks where crack masks are overlaid on the texture of masonry buildings with the appropriate parameters set as input. 
    \item We proposed an evaluation method based on how much the real dataset can be decreased while augmenting the synthetic dataset to enhance the crack detection results of the CNN model considered. 
\end{itemize}
The remain works of this paper are as follows: Section 2 introduces the related works on DL crack detection models and synthetic crack generation methods. In Section 3 we discuss the pipeline of the proposed methodology. 
Section 4 provides a performance evaluation of the models involved in crack detection. Finally, the conclusion and future works are presented in Section 5.

\section{Related Works}
Crack detection based on DL algorithms is a spotlight in the field of structural monitoring since it provides effortless solutions with respect to more traditional approaches \cite{mfafnet_dong_2024}, \cite{deep_katsigiannis_2023}. 
In particular, different DL methods were used and compared in order to find the best output for both bricks and crack segmentation, demonstrating great potential in the documentation of masonry fabrics \cite{automatic_loverdos_2022}. 
However, most of these methods still suffer the common problem of limited images availability, prompting researchers to develop data augmentation methods based on synthetic data generation. In this scenario, a powerful approach is represented by crack images synthesized from labeled images using a conditional GAN to enhance data diversity in the learning process \cite{self_shim_2024}. Marsh et al. \cite{crack_marsh_2025} trained an existing architecture (DeepLabV3) with both real-world data and synthetic data generated with Blender. Chu et al. \cite{crackgaugan_chu_2024} proposed CrackGauGAN, a network for automated realistic crack images and masks generation, where a Criminisi-based crack image operator is added to exclude crack noise interference and a background extractor is proposed as texture decoupling. To reduce the workload in DL-based detection, Li et al. \cite{data_li_2024} proposed two improvements to CycleCANs (Tiny-CycleGAN and Multi-CycleGAN) with the aim of automatizing annotation and enhancing the real dataset. Dondi et al. \cite{improving_dondi_2025} demonstrated that methods based on a combination of real and semi-synthetic images outperform systems trained with only real images. In particular, they employed parametric meta-annotations to generate cracks on 3D models of real structures. Unlike those approaches, our framework generates synthetic crack with a simplified cracks overlaying tool (allowing for a controlled insertion of cracks in the masonry wall background) and tests the performance of a balanced combination of real/synthetic dataset to determine the minimum amount of real images to improve the overall results.   

\section{Methodology}
\begin{figure}[htbp!]
    \centering
    \includegraphics[width=1\linewidth]{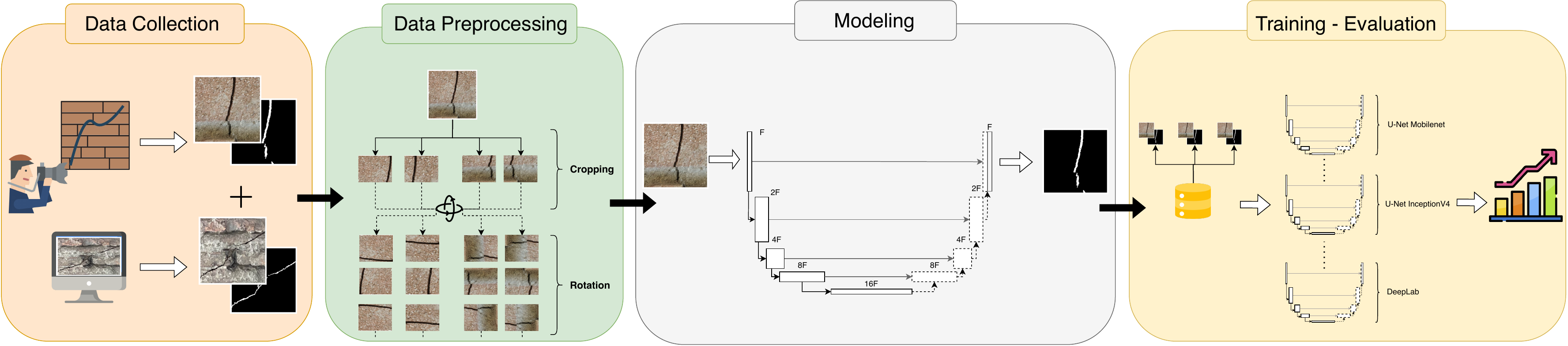}
    \caption{Overview of the proposed pipeline for masonry crack segmentation, from data collection to preprocessing, augmentation and modeling}
    \label{fig:pipeline}
\end{figure}

This section presents the implemented image segmentation pipeline, shown in Figure \ref{fig:pipeline}, consisting of four stages:
data acquisition, data preprocessing, model architecture and experimental setup. The pipeline starts with the acquisition of crack images from masonry surfaces. 
In addition, a process was employed to generate synthetic images and increase the variability of crack patterns representative of real masonry surfaces (\textit{data collection}). Once the dataset was collected and annotated, to improve model generalization, data processing and data augmentation techniques were applied (\textit{data preprocessing}).
Finally, several SOTA U-Net architectures (\textit{model architecture}) were then employed to enable a systematic comparison of our approach with current methods in the literature (\textit{performance evaluation}). 
This section is structured as follows: Subsection \ref{subs:datacollection}  details the acquisition methodology, devices used and criteria applied to create a reliable dataset for crack segmentation. Subsection \ref{subs:dataprocessing} details the preprocessing steps. Subsection \ref{subs:modeling} introduces the modeling choices, with a focus on U-Net variants. Finally, Subsection \ref{subs:experimental} outlines the experimental setup, covering training strategies and evaluation metrics.

\subsection{Data Collection}\label{subs:datacollection}
Two different datasets were collected: (\textit{i}) real-world dataset representing real masonry cracks, (\textit{ii}) synthetic dataset generated from real masonry surfaces.
Images were acquired by three volunteers at multiple outdoor locations, using standard smartphones. The usage of common smartphones (Oppo Reno 6 pro and Iphone 12 \& 13)
make the collection process both cost-effective and easily reproducible. 
    
All pictures were captured during daylight to ensure adequate illumination, with the camera positioned as orthogonally as possible to the wall surface. 
An example of a crack image is depicted in Figure \ref{fig:preprocessing}.

\begin{figure}[htbp!]
    \centering
        \subfigure{
          \includegraphics[width=.55\linewidth]{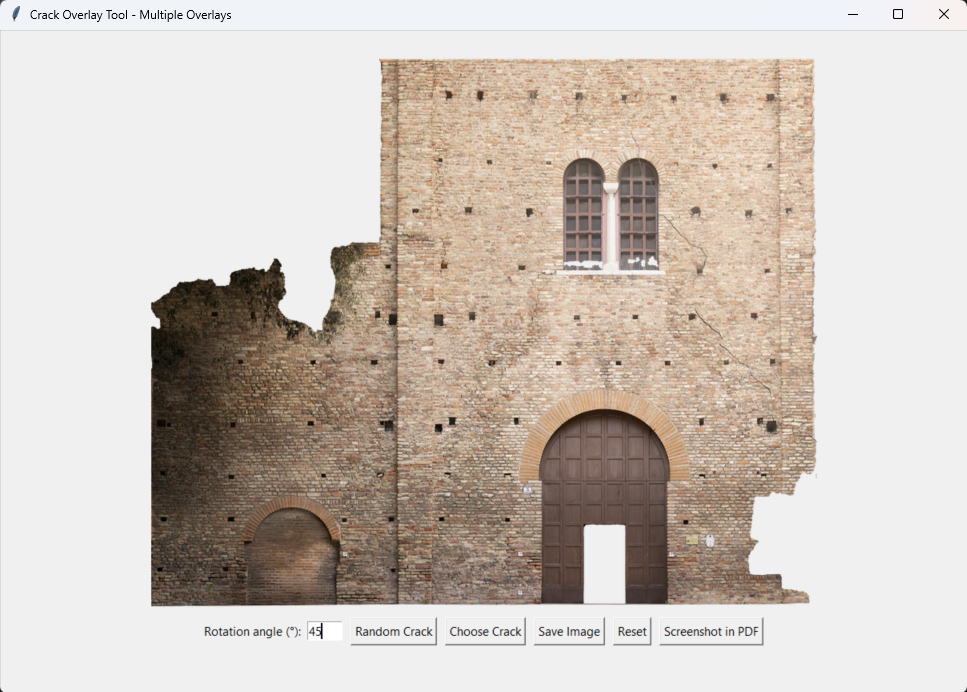}
            \label{fig:tool}
        }
        \subfigure{
            \includegraphics[width=.4\linewidth]{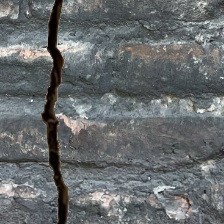}
            \label{fig:crack_img}
        }
        \caption{(a) Masonry wall of a church in Ravenna. (b) Crack image generated. }
        \label{fig:gen_cracks}
\end{figure}

The next step in the data collection process was the generation of a synthetic dataset to address the limited availability of real crack images. For this purpose, a simple custom-made software tool was developed to simulate cracks on masonry surfaces. 
The tool is a Python-based application implementing an interactive graphical user interface using Tkinter for the visual annotation of structural damage. 
It enables users to overlay multiple semi-transparent crack images onto a high-resolution background (e.g., architectural or orthoimagery), with support for spatial placement and arbitrary rotation. The tool preserves scale consistency by adjusting overlays according to the resized display while maintaining export fidelity on the original-resolution image. The final composition can be saved as a high-resolution PNG and added to the dataset. An example of the wall surface used for the data generation and a crack generated are shown in the Figure \ref{fig:gen_cracks}.

\subsection{Data preprocessing}\label{subs:dataprocessing}
\begin{figure}[ht]
    \centering
    \includegraphics[width=0.67\linewidth]{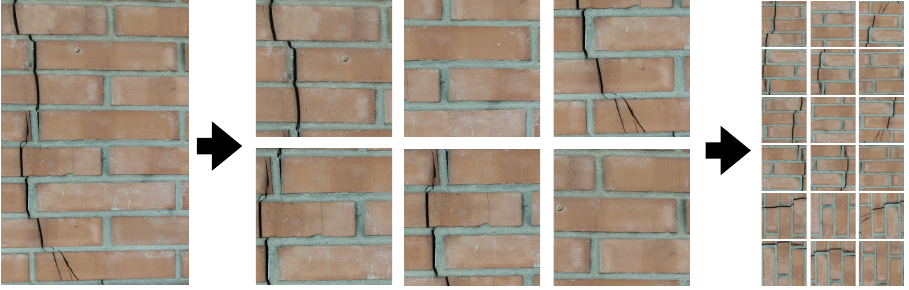}
    \caption{Example of preprocessing pipeline: raw image (left), cropped patch (center), and augmented variants (right)}
    \label{fig:preprocessing}
\end{figure}

After collecting the datasets, the raw images were preprocessed through four steps. First, images were cropped into 224×224 patches to standardize resolution across devices and increase crack samples. Second, ground-truth masks were created via pixel-wise annotation (crack vs. background). Third, data augmentation was applied using flips and 90° rotations to further expand the dataset. An example of these preprocessing steps is shown in Figure \ref{fig:preprocessing}.
The augmented images were then used for training the model only, while for evaluation, only the real patches were used. Finally, the images were normalized according to ImageNet statistics to ensure compatibility with the pre-trained U-Net backbones used in this study.
\subsection{Model Architecture}\label{subs:modeling}
The U-Net architecture was employed. It follows an encoder–decoder design: the encoder reduces spatial resolution to capture contextual information, while the decoder restores details through upsampling. Skip connections between the encoder and decoder layers combine semantic and spatial features, making the U-Net architecture highly effective for pixel-level segmentation. 
To evaluate the performance of different backbone encoders, several SOTA CNNs were employed within the U-Net architecture. 
In addition to the SOTA backbones, we implemented a custom U-Net in Pytorch to serve as a baseline. The model follows the canonical encoder-decoder design described before. 

In Table \ref{tab:unets} are shown the models used to evaluate the segmentation task. 
\begin{table}[ht]
    \centering
    \caption{Summary of the backbone architectures used in this study. All models were initialized with pretrained weights.}
    \begin{tabular}{lccc}
        \toprule
        \textbf{Model} & \textbf{Parameters} & \textbf{Size (MB)} & \textbf{Pretrained} \\
        \midrule
        Custom U-Net        & 36,963,266 & 147.85 & No \\
        VGG16 (U-Net) \cite{simonyan2014very}       & 23,748,386 & 94.99 & Yes \\
        MobileNetV2 (U-Net) \cite{sandler2018mobilenetv2} & 6,629,090 & 26.52 & Yes \\
        ResNet50 (U-Net) \cite{he2016deep}    & 32,521,250 & 130.09 & Yes \\
        DenseNet169 (U-Net) \cite{huang2017densely} & 21,202,786 & 84.81 & Yes \\
        InceptionV4 (U-Net) \cite{szegedy2017inception} & 48,792,066 & 195.17 & Yes \\
        DeepLabV3 (ResNet50) \cite{chen2017deeplab} & 26,398,295 & 105.59 & Yes \\
        \bottomrule
    \end{tabular}
    
    \label{tab:unets}
\end{table}

\subsection{Experimental Setup}\label{subs:experimental}
The experimental setup was designed to ensure a fair evaluation of different model-loss configurations under same conditions (real images only). The real image dataset was partitioned into training, validation and test sets with a 60/20/20 ratio. Synthetic samples were introduced only into the training set of the best performing model, while validation and test were preserved, as shown in Table \ref{tab:inception_results}. Three tasks-specific loss functions were used to train each architecture: \textit{(i)} F1Loss (F1), which aims to optimize the segmentation quality balancing precision and recall per-class, \textit{(ii)} FocalLoss (FL) \cite{lin2017focal}, which focus on hard-to-classify pixels and \textit{(iii)} a combination of CrossEntropy \cite{bishop2006pattern} and Dice Loss \cite{milletari2016vnet} (CED) aiming to balance region overlap with pixel-wise accuracy. The adoption of these losses is consistent with the SOTA and has been employed in related studies, including \cite{automatic_dais_2021}.
The training was performed with a maximum of 100 epochs. Early stopping criteria was applied if no improvement in validation performance was observed for 10 consecutive epochs, promoting convergence while preventing overfitting. Table \ref{tab:inception_results} shows the number of real and synthetic images used to train the models. This figure does not take into account the augmentation performed during the training process.


\begin{table}[ht]
    \centering
    \footnotesize
    \caption{Results of InceptionV4 using different training set composition.}
    \begin{tabular}{lcc}
        \toprule
        \makecell[l]{\textbf{Training set} \\ \textbf{(\%syn/\%real: n. real + n. syn)}} & \textbf{Validation set} & \textbf{Test set} \\
        \midrule
        100/0: 504 syn + 0 real        & 139 real & 140 real  \\
        80/20: 403 syn + 83 real       & 139 real & 140 real \\
        60/40: 302 syn + 166 real      & 139 real & 140 real \\
        40/60: 201 syn + 250 real      & 139 real & 140 real \\
        20/80: 100 syn + 333 real      & 139 real & 140 real  \\
        100/0: 0 syn + 417 real        & 139 real & 140 real  \\
        \bottomrule
    \end{tabular}
    \label{tab:inception_results}
\end{table}

\section{Performance Evaluation}\label{sez:performance_evaluation}

\begin{figure}[ht]
    \centering
    \includegraphics[width=0.9\linewidth]{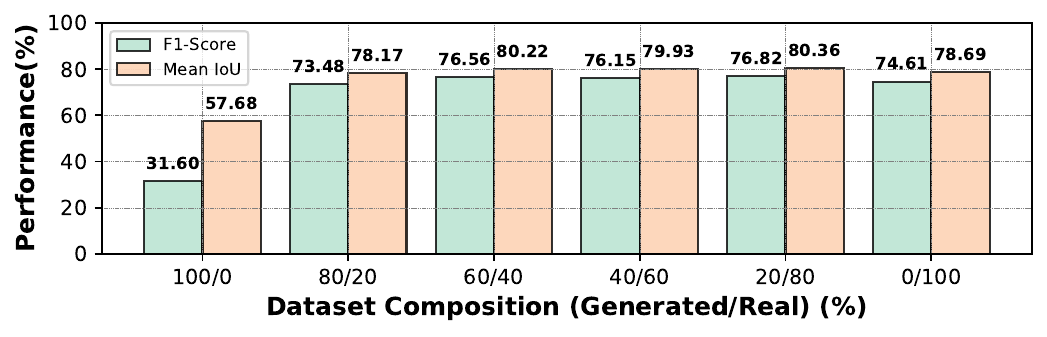}
    \caption{InceptionV4 F1-score results for different compositions of generated/real cracks  (training datasets) and real-only cracks (test dataset).}
    \label{fig:f1_balance_dataset}
\end{figure}

The performance of the proposed framework is evaluated in two phases. Initially, U-Net architectures were trained with only real image data to establish the best-performing model. Then, the selected model was trained with different combinations of synthetic/real images to determine the minimum contribution of real images necessary to reach results similar or even higher than those related to real data scenario. 
Table \ref{tab:results_f1_miou} shows the prediction results in terms of F1-score and mIoU achieved by the different loss functions used for the models analyzed.
The diagram exhibits outstanding and similar segmentation results from many recent SOTA methods in the field of crack detection of masonry structures. In particular, each loss function exceeds the 70\% F1-score and mean IoU, with the custom U-Net achieving a comparable result with respect to the other models except for CED where the maximum F1-score reached is 65\%. The best analyzed model is InceptionV4 (74.7\% F1-score, 78\% mean IoU) which was chosen for experiments with synthetic data and as a reference case. 
Figure \ref{fig:f1_balance_dataset} shows F1-score and mean IoU results for six balanced combinations of generated and real datasets used to train InceptionV4. Since the model was tested with the real dataset only, the different train scenarios produce different results that were compared with the reference case. As expected, the first scenario (100\% synthetic dataset) is the one that performs the worst, since the model was trained and tested with two different types of images (synthetic and real). Conversely we noticed that reducing the generated synthetic data and proportionally increasing the real images counterpart, the F1 metric tends to increase as well reaching the maximum results (80.3\%) at the 20/80 scenario. The overall analysis demonstrates the potentiality of the method in reaching even better results considering a smaller number of real images. Another important achievement is represented by the case 80/20 where reducing at 20\% the real dataset and increasing at 80\% the synthetic data the model can reach almost identical results of the scenario with only real images. Hence, the method demonstrates the capability to reach outstanding results, minimizing the contribution of real data and maximizing the contribution of synthetic data.

        
        

\begin{table}[t]
    \centering
     \caption{Comparison across models using three losses. Metrics reported as percentages.}
    \label{tab:results_f1_miou}
    \small
    \begin{tabular}{lcccccc}
        \toprule
        \multirow{2}{*}{Model} & \multicolumn{3}{c}{F1 Score (\%)} & \multicolumn{3}{c}{Mean IoU (\%)} \\
        \cmidrule(lr){2-4}\cmidrule(lr){5-7}
         & CED Loss & F1-Loss & Focal Loss & CED Loss & F1-Loss & Focal Loss \\
        \midrule
        Custom U-Net         & 65.51 & 72.21 & 69.68 & 72.57 & 77.11 & 75.24 \\
        VGG16        & 70.84 & 73.17 & 72.90 & 76.02 & 77.76 & 77.38 \\
        MobileNetV2  & 72.67 & 73.86 & 72.75 & 77.22 & 78.18 & 77.31 \\
        ResNet50     & 72.69 & 72.27 & 72.63 & 77.22 & 77.13 & 77.22 \\
        DenseNet169  & 74.13 & 72.20 & 73.37 & 78.26 & 77.09 & 77.72 \\
        \textbf{InceptionV4}  & \textbf{74.70} & \textbf{74.61} & \textbf{73.94} & \textbf{78.65} & \textbf{78.69} & \textbf{78.13} \\
        DeepLabv3    & 70.42 & 70.10 & 71.53 & 75.72 & 75.71 & 76.49 \\
        \bottomrule
    \end{tabular}
   
\end{table}

\section{Conclusion and Future Works}

In this paper, we proposed and investigated methods for detecting cracks in masonry images with the enhancement of synthetic crack generation. The research has initially focused on selecting the best CNN model for crack detection task of real masonry crack images collected by our laboratory. 
Secondly, we proposed a custom image-processing-based synthetic data generator to improve the results obtained for the best model (InceptionV4: 74.7\% F1-score). The simplicity of the tool, which relies on crack's mask overlay, is conceived to minimize the effort in data production while achieving improved segmentation performance. The method is thought to discern those combinations of generated/real training images that could positively influence the models segmentation tested with real data. Two training scenarios were considered of particular relevance: the 80/20 (73\% F1-score, 78\% mean IoU) and the 20/80 (76\% F1-score, 80.3\% mean IoU). 
The two results demonstrate that with a very simple crack generation overlay tool, the performances increase by 2\% for the 20/80 scenario, while reaching the same reference result (0/100: 74.6\% F1-score, 78.6\% mean IoU) in the 80/20 scenario.
In the future, this approach will be extended to compare the result of the custom overlay method with other SOTA methods (such as GANs and 3D computer graphic software) in order to improve crack predictions. Moreover, the framework will be extended to quantify crack features like width and length, making it a cost-effective and practical alternative to traditional detection and monitoring methods.
\bibliographystyle{unsrt}
\bibliography{bibliography}
%

\end{document}